\documentclass[journal,twoside,web]{ieeecolor}
\usepackage{tmi}
\usepackage{cite}
\usepackage{amsmath,amssymb,amsfonts}
\usepackage{algorithmic}
\usepackage{graphicx}
\usepackage{mathrsfs}
\usepackage{subfig}
\usepackage{textcomp}
\usepackage{tablefootnote}
\def\BibTeX{{\rm B\kern-.05em{\sc i\kern-.025em b}\kern-.08em
    T\kern-.1667em\lower.7ex\hbox{E}\kern-.125emX}}
\begin{document}
\title{Two-Stage Mesh Deep Learning for Automated Tooth Segmentation and Landmark Localization on 3D Intraoral Scans}
\author{Tai-Hsien Wu, Chunfeng Lian, \IEEEmembership{Member, IEEE}, Sanghee Lee, Matthew Pastewait, Christian Piers, Jie Liu, Fan Wang, Li Wang, \IEEEmembership{Senior Member, IEEE}, Chiung-Ying Chiu, Wenchi Wang, Christina Jackson, Wei-Lun Chao, Dinggang Shen, \IEEEmembership{Fellow, IEEE}, and Ching-Chang Ko
\thanks{The manuscript was submitted on 04/08/2021.}
%``\textcolor[rgb]{0,0,0}{This work was supported in part by the U.S. Department of Commerce under Grant BS123456.}''
\thanks{T. Wu and C. Lian contributed equally to this work.}
\thanks{Corresponding authors: C. Ko (ko.367@osu.edu) and D. Shen (dinggang.shen@gmail.com)}
\thanks{T. Wu, S. Lee, J. Liu, and C. Ko are with the Division of Orthodontics, College of Dentistry, the Ohio State University, Columbus 43210 ,USA.}
\thanks{C. Lian is with the School of Mathematics and Statistics, Xi'an Jiaotong University, Xi'an, Shaanxi 710049, China.}
\thanks{M. Pastewait is with the United States Air Force, Kadena AB, Japan.}
\thanks{C. Piers is in private practice in Morganton, NC, 286555, USA.}
\thanks{F. Wang is with the Key Laboratory of Biomedical Information Engineering of Ministry of Education, Department of Biomedical Engineering, School of Life Science and Technology, Xi'an Jiaotong University, Xi'an, Shaanxi, 710049, China.}
\thanks{L. Wang is with the Department of Radiology and BRIC, University of North Carolina at Chapel Hill, Chapel Hill, NC 27599, USA.}
\thanks{C. Chiu, W. Wang, and C. Jackson are with the SOVE Inc, Columbus 43220, USA.}
\thanks{W. Chao is with the Department of Computer Science and Engineering, the Ohio State University, Columbus 43210, USA.}
\thanks{D. Shen is with the School of Biomedical Engineering, ShanghaiTech University Shanghai, China, and Department of Research and Development, Shanghai United Imaging Intelligence Co., Ltd., Shanghai, China.}
}

\maketitle

\begin{abstract}
Accurately segmenting teeth and identifying the corresponding anatomical landmarks on dental mesh models are essential in computer-aided orthodontic treatment.
Manually performing these two tasks is time-consuming, tedious, and, more importantly, highly dependent on orthodontists' experiences due to the abnormality and large-scale variance of patients' teeth.
Some machine learning-based methods have been designed and applied in the orthodontic field to automatically segment dental meshes (e.g., intraoral scans).
In contrast, the number of studies on tooth landmark localization is still limited.
This paper proposes a two-stage framework based on mesh deep learning (called TS-MDL) for joint tooth labeling and landmark identification on raw intraoral scans.
Our TS-MDL first adopts an end-to-end \emph{i}MeshSegNet method (i.e., a variant of the existing MeshSegNet with both improved accuracy and efficiency) to label each tooth on the downsampled scan.
Guided by the segmentation outputs, our TS-MDL further selects each tooth's region of interest (ROI) on the original mesh to construct a light-weight variant of the pioneering PointNet (i.e., PointNet-Reg) for regressing the corresponding landmark heatmaps.
Our TS-MDL was evaluated on a real-clinical dataset, showing promising segmentation and localization performance.
Specifically, \emph{i}MeshSegNet in the first stage of TS-MDL reached an averaged Dice similarity coefficient (DSC) at \textcolor[rgb]{0,0,0}{$0.964\pm0.054$}, significantly outperforming the original MeshSegNet.
In the second stage, PointNet-Reg achieved a mean absolute error (MAE) of \textcolor[rgb]{0,0,0}{$0.597\pm0.761 \, mm$} in distances between the prediction and ground truth for \textcolor[rgb]{0,0,0}{$66$} landmarks, which is superior compared with other networks for landmark detection.
All these results suggest the potential usage of our TS-MDL in \textcolor[rgb]{0,0,0}{orthodontics}.
\end{abstract}

\begin{IEEEkeywords}
Tooth Segmentation, Anatomical Landmark Detection, Orthodontic Treatment Planning, 3D Deep Learning, Intraoral Scan.
\end{IEEEkeywords}

\section{Introduction}
\label{sec:introduction}
Digital 3D dental models have been widely used in orthodontics due to their efficiency and safety.
To create a patient-specific treatment plan (e.g., for the fabrication of clear aligners), orthodontists need to segment teeth and annotate the corresponding anatomical landmarks on 3D dental models to analyze and rearrange tooth positions.
Manually performing these two tasks is time-consuming, tedious, and expertise-dependent, even with the assistance provided by most commercial software (integrating semi-automatic algorithms).
Although there is a clinical need to develop fully automatic methods instead of manual operations, it is practically challenging, especially for tooth landmark localization, mainly due to (1) large-scale shape variance of different teeth, (2) abnormal, disarranged, and/or missing teeth for some patients,  and (3) incomplete dental models (e.g., the second/third molars) captured by intraoral scanners.

\textcolor[rgb]{0,0,0}{Almost 15 year ago, Zhao et al. used curvature values of mesh cells to generate feature contours for tooth segmentation in a semi-automatic manner, indicating the importance of segmentation on dental mesh models in modern dentistry~\cite{Zhao2005}.}
Recently, some deep learning approaches~\cite{Xu2019, Tian2019, pmlr-v102-ghazvinian-zanjani19a, Sun2020, Zhang2020, Zanjani2019, Cui2021} have been proposed for end-to-end dental surface labeling.
Although these deep networks show superior segmentation accuracy than conventional (semi-)automatic methods, mainly due to task-oriented extraction and fusion of local details and semantic information, very few of them address the critical task of landmark localization.
Compared with tooth segmentation, localizing anatomical landmarks is typically more sensitive to varying shape appearance of patients' teeth, as each tooth's landmarks are just small points encoding local geometric details, and the number of landmarks changes across positions.

\begin{figure}[!t]
\centerline{\includegraphics[width=\columnwidth]{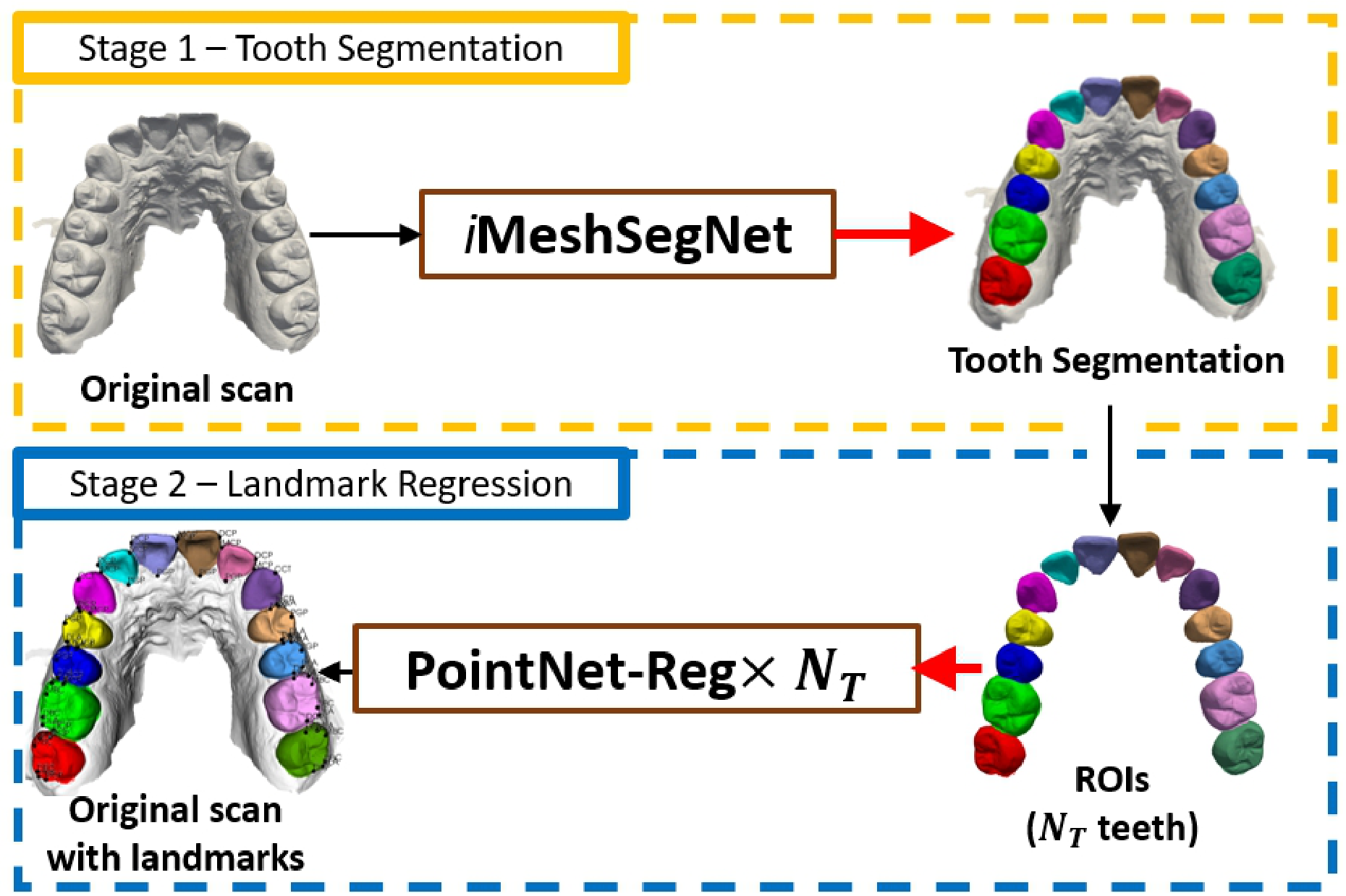}}
\caption{The workflow of our two-stage method for automated tooth segmentation and dental landmark localization.}
\label{fig_TS-MDL}
\end{figure}

Considering the natural correlations between the two tasks (e.g., each tooth's landmarks depend primarily on its local geometry), a two-stage framework leveraging mesh deep learning (called TS-MDL) is proposed in this paper for joint tooth segmentation and landmark localization.
The schematic diagram of our TS-MDL is shown in Fig.~\ref{fig_TS-MDL}.

In Stage 1, we propose an end-to-end deep neural network, called \emph{i}MeshSegNet, to perform tooth segmentation on 3D intraoral scans. \emph{i}MeshSegNet improves the implementation of the multi-scale graph-constrained learning module in its forerunner MeshSegNet~\cite{Lian2020, Lian2019}, a neural network for automatic tooth segmentation.
In Stage 2, we extract cells that belong to individual teeth based on the segmentation results generated by Stage 1. By doing this, we narrow the entire intraoral scan down to several ROIs (i.e., individual teeth) since a tooth landmark is always on and only associated with its corresponding tooth.
Inspired by the use of heatmaps to successfully detect anatomical landmarks on 2D and 3D medical images~\cite{Payer2016}, we design a modified PointNet~\cite{Qi2016}, called PointNet-Reg, to learn the heatmaps encoding landmark locations.
The experimental results on real clinical data indicate that \emph{i}MeshSegNet improves tooth segmentation in terms of both accuracy and efficiency, and the straightforward two-stage strategy leads to promising accuracy in anatomical landmark localization.

The rest of the paper is organized as follows.
We briefly review the most related work in Section II, including deep learning on 3D dental models for automated tooth segmentation and heatmap regression for anatomical landmark localization in medical images.
Section III describes the studied data and our TS-MDL method.
The experimental results and the comparisons between our method and other strategies/approaches are presented in Section IV.
We further discuss the effectiveness of some critical methodological designs and our current method's potential limitations in Section V.
Finally, the work is concluded in Section VI.

\section{Related Work}
\subsection{Deep Learning-based 3D Dental Mesh Labeling}
Several deep learning-based methods, leveraging convolutional neural networks (CNNs) or graph neural networks (GNNs), have been proposed for automated tooth segmentation on 3D dental meshes.
For the CNN-based method, Xu et al.~\cite{Xu2019} extracted cell-level hand-crafted features to form 2D image-like inputs of a CNN, which predicts the respective semantic label of each cell.
Tian et al.~\cite{Tian2019} converted the dental mesh to a sparse voxel octree model and used 3D CNNs to segment and identify individual teeth.
Zhang et al.~\cite{Zhang2020} mapped a 3D tooth model isomorphically to a 2D "image" encoding harmonic attributes. A CNN model \textcolor[rgb]{0,0,0}{was} further learned to predict the segmentation mask, which \textcolor[rgb]{0,0,0}{was} then transferred back to the original 3D space.
These CNN-based methods cannot work directly on the raw dental surfaces, and they typically need to convert 3D meshes (or point clouds) to regular "images" (of hand-crafted features), inevitably losing fine-grained geometric information of teeth.

Inspired by the pioneering PointNet~\cite{Qi2016} working directly on the 3D shapes for shape-level or point/cell-level classifications, an increasing number of studies proposed to design GNN-based methods for mesh or point cloud segmentation in an end-to-end fashion.
For example, Zanjani et al.~\cite{pmlr-v102-ghazvinian-zanjani19a} combined PointCNN~\cite{Li2018} with a discriminator in an adversarial setting to automatically assign tooth labels to each point from intraoral scans.
%Sun et al.~\cite{Sun2020} used FeaStNet~\cite{Verma2018} to automatically label teeth on digital dental casts.
Sun et al.~\cite{Sun2020} proposed to label teeth on digital dental casts by using FeaStNet~\cite{Verma2018}, which was further extended in their more recent work~\cite{Sun2020a} for coupled tooth segmentation and landmark localization.
\textcolor[rgb]{0,0,0}{Clinically, the number of teeth can vary between patients, which increases the difficulty of the segmentation task. In order to address this issue, Zanjani et al. proposed Mask-MCNet, which is analogous to Mask-RCNN~\cite{He2017b}, to conduct instance segmentation on intra-oral scans~\cite{Zanjani2019}. In addition, Cui et al. proposed TSegNet, which detects all the teeth using a distance-aware tooth centroid voting scheme, followed by a confidence-aware cascade segmentation module to outperform state-of-the-art approaches~\cite{Cui2021}.}

Our previous work, MeshSegNet~\cite{Lian2020, Lian2019}, is also based on GNN, which integrates a series of graph-constrained learning modules to hierarchically extract and integrate multi-scale local contextual features for tooth labeling on raw intraoral scans.
Although it has achieved state-of-the-art segmentation accuracy, MeshSegNet has a key limitation -- the heavy \textcolor[rgb]{0,0,0}{computational requirements} due to the large-scale matrix computations of the large adjacent matrices.

%\subsection{Heatmap Regression-based Landmark Localization}
\subsection{\textcolor[rgb]{0,0,0}{Learning-based Landmark Localization}}
Landmark localization is a crucial task in \textcolor[rgb]{0,0,0}{both computer vision and} medical imaging analysis\textcolor[rgb]{0,0,0}{.}
\textcolor[rgb]{0,0,0}{In 2012, Kumer et al. presented a set of specific methods to automatically identify several dental-specific features (e.g., cusps, marginal ridges, grooves, etc.) on digital dental meshes~\cite{Kumar2012}. However, each method is specific to identify the corresponding feature only, which means this system cannot detect landmarks outside their well-defined domain.}
\textcolor[rgb]{0,0,0}{In deep learning, a} straightforward strategy for landmark localization is to regress the coordinates directly from high-dimensional inputs (e.g., images), but learning such highly nonlinear mappings is technically challenging~\cite{Pfister2015}.
To deal with this challenge, various studies in the computer vision community (e.g., Pfister et al.~\cite{Pfister2015}) proposed to encode the location information of landmarks into Gaussian heatmaps, by which the point localization task is transferred as an easier image-to-image/heatmap regression task.
\textcolor[rgb]{0,0,0}{The task of landmark localization is still a very active topic in computer vision, and several novel studies recently emerged with a focus on self-supervised methods.}
\textcolor[rgb]{0,0,0}{For example, Suwajanakorn et al. presented an end-to-end geometric reasoning framework to learn a latent set of category-specific 2D keypoints with depth information, optimized by multi-view consistency and relative pose estimation~\cite{Suwajanakorn2018}. Reddy et al. also presented a graph-based framework, called Occlusion-Net, with a largely self-supervised scheme to predict the 2D and 3D locations of occluded keypoints~\cite{Reddy2019}.}

\textcolor[rgb]{0,0,0}{In medical imaging analysis, the landmarking task} has been widely used in diseases diagnosis, treatment planning, and surgical simulation~\cite{Zhang2016,Meng2014,Zhang2017,Lian2020a}.
The heatmap regression-based landmark localization has also been successfully applied in the medical imaging domain.
For example, by leveraging the end-to-end image-to-image learning ability of fully convolutional networks (FCNs), Payer et al. proposed SpatialConfiguration-Net that incorporates spatial configuration of anatomical landmarks with local appearance to improve \textcolor[rgb]{0,0,0}{the} robustness of heatmap regression even with a limited amount of training data~\cite{Payer2016, Payer2019}.
In addition, due to the association between segmentation and landmarking, Zhang et al. \cite{Zhang2020a} aimed to solve the two tasks jointly and adopted heatmap regression for landmark digitization in their approach.
In general, these existing studies only focus on detecting landmarks in 2D/3D medical images, without attention to more complicated data structures such as irregular dental meshes.

\textcolor[rgb]{0,0,0}{Recently, some studies have started working on landmark identification for point cloud data. For instance, Fernandez-Labrador et al. proposed an unsupervised learning method for category-specific keypoint identification on 3D point clouds of objects~\cite{Fernandez-Labrador2020}.}
\textcolor[rgb]{0,0,0}{Maron et al. proposed a method to transfer the surface mesh to a flat torus (i.e., a 2D image) so that the traditional CNN would be able to apply to the corresponding flat torus~\cite{Maron2017}.}
\textcolor[rgb]{0,0,0}{However, the number of studies regarding landmark detection on dental mesh models\textcolor[rgb]{0,0,0}{, particularly directly working on 3D data with graph neural networks,} is still limited.}

\section{Materials and Method}
\subsection{Dataset}
\begin{figure}[!t]
\centerline{\includegraphics[width=\columnwidth]{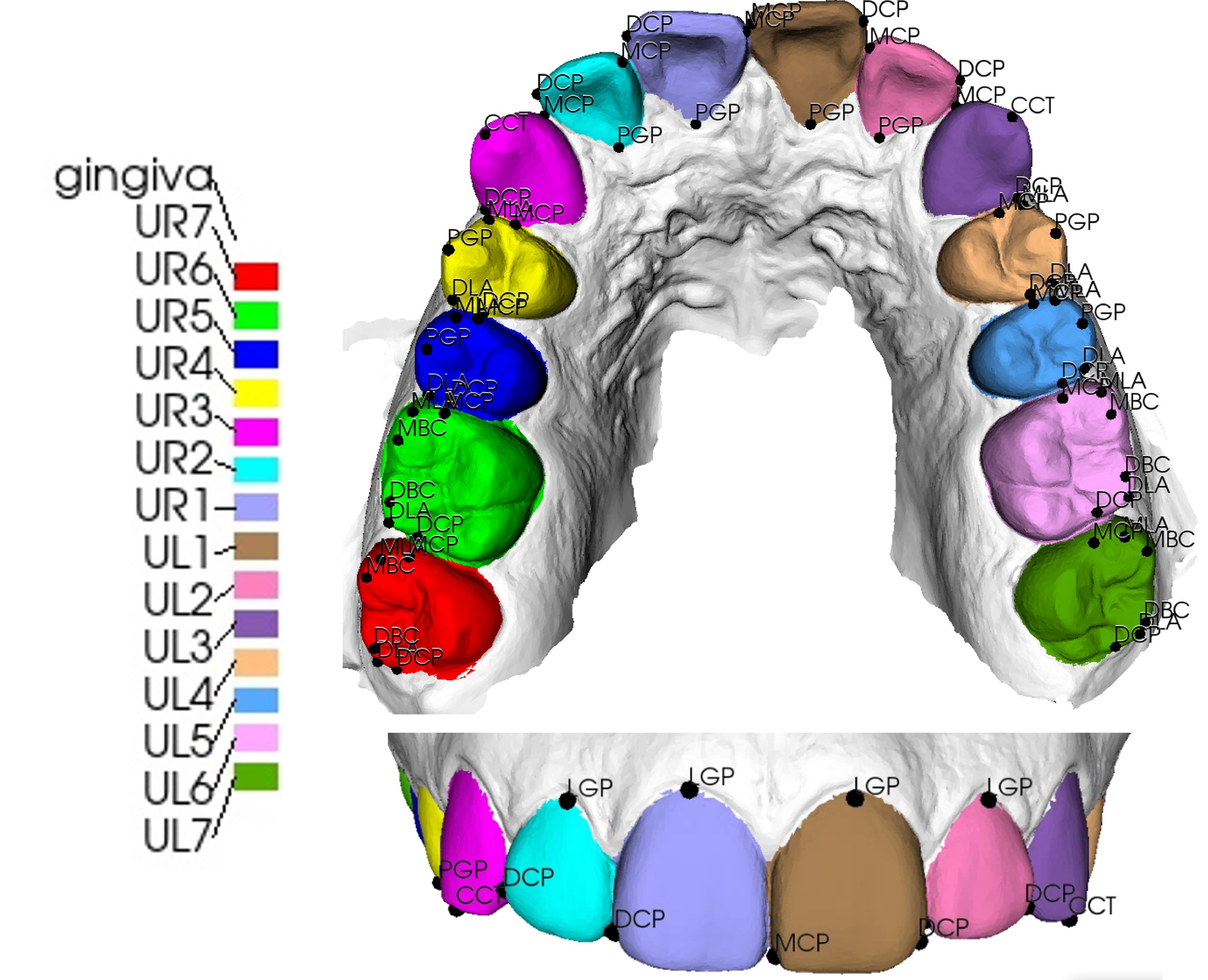}}
\caption{The learning targets in this study are the 14 teeth and \textcolor[rgb]{0,0,0}{66} anatomical landmarks on \textcolor[rgb]{0,0,0}{14} teeth. The top and bottom graphics are occlusal and \textcolor[rgb]{0,0,0}{frontal} views, respectively.}
\label{fig_44_landmarks}
\end{figure}

The dataset used in this study consists of \textcolor[rgb]{0,0,0}{136} patients' raw upper intraoral scans (mesh surfaces), acquired by iTero Element\textregistered, a 3D dental intraoral scanner (IOS).
\textcolor[rgb]{0,0,0}{The Institutional Review Boards (IRBs) used in this study are IRB\# $13$-$0924$ (University of North Carolina at Chapel Hill) and 2020H0459 (The Ohio State University).}
With the exception of a few cases with missing teeth, each scan has 14 teeth and \textcolor[rgb]{0,0,0}{66} landmarks.
\textcolor[rgb]{0,0,0}{These 66 landmarks are commonly recognized in the orthodontic field and are useful in superimposition as well as calculating tooth movement between pre- and post-treatment scans. The full name for each landmark acronym is given in Table~\ref{table_1}.}
Each scan was manually annotated and checked by two experienced orthodontists, serving as the ground truth for the segmentation and localization tasks, respectively.
A typical example is shown in Fig.~\ref{fig_44_landmarks}, with the names of teeth and corresponding landmarks listed in Table~\ref{table_1}.

\begin{table}[!t]
\caption{The names of landmarks, numbers of landmarks, and their corresponding teeth.}
\begin{tabular}{llc}
\hline
\textbf{Tooth name}                                                          & \textbf{Landmark name} & \multicolumn{1}{l}{\textbf{No. of landmarks}} \\ \hline
\begin{tabular}[c]{@{}l@{}}upper right/left \textcolor[rgb]{0,0,0}{central} incisor\\ (UR1, UL1)\end{tabular}         & \begin{tabular}[c]{@{}l@{}}DCP, MCP, PGP,\\ LGP\end{tabular}           & 4 \\
\begin{tabular}[c]{@{}l@{}}upper right/left lateral incisor\\ (UR2, UL2)\end{tabular} & \begin{tabular}[c]{@{}l@{}}DCP, MCP, PGP,\\ LGP\end{tabular}           & 4 \\
\begin{tabular}[c]{@{}l@{}}upper right/left canine\\ (UR3, UL3)\end{tabular} & DCP, MCP, CCT          & 3                                                \\
\begin{tabular}[c]{@{}l@{}}upper right/left first premolar\\ (UR4, UL4)\end{tabular}  & \begin{tabular}[c]{@{}l@{}}MLA, DLA, PGP,\\ MCP, DCP\end{tabular}      & 5 \\
\begin{tabular}[c]{@{}l@{}}\textcolor[rgb]{0,0,0}{upper right/left second premolar}\\ \textcolor[rgb]{0,0,0}{(UR5, UL5)}\end{tabular}  & \begin{tabular}[c]{@{}l@{}}\textcolor[rgb]{0,0,0}{MLA, DLA, PGP,}\\ \textcolor[rgb]{0,0,0}{MCP, DCP}\end{tabular}      & \textcolor[rgb]{0,0,0}{5} \\
\begin{tabular}[c]{@{}l@{}}upper right/left first molar\\ (UR6, UL6)\end{tabular}     & \begin{tabular}[c]{@{}l@{}}MLA, DLA, MBC,\\ DBC, MCP, DCP\end{tabular} & 6 \\
\begin{tabular}[c]{@{}l@{}}\textcolor[rgb]{0,0,0}{upper right/left second molar}\\ \textcolor[rgb]{0,0,0}{(UR7, UL7)}\end{tabular}     & \begin{tabular}[c]{@{}l@{}}\textcolor[rgb]{0,0,0}{MLA, DLA, MBC,}\\ \textcolor[rgb]{0,0,0}{DBC, MCP, DCP}\end{tabular} & \textcolor[rgb]{0,0,0}{6} \\ \hline
\end{tabular}
\footnotesize{\textcolor[rgb]{0,0,0}{MCP (mesial contact point); DCP (distal contact point);\\PGP (palatal gingival point); LGP (labial gingival point);\\CCT (canine cusp tip);\\MLA (mesial line angle); DLA (distal line angle);\\MBC (mesiobuccal cusp); DBC (distobuccal cusp)}}\\
\label{table_1}
\end{table}

\subsection{TS-MDL}
In this paper, we propose a two-stage method for automated identification of anatomical landmarks on 3D intraoral scans.
The workflow of this method is shown in Fig.~\ref{fig_TS-MDL}, where the two stages correspond to tooth segmentation and heatmap regression, respectively.
The idea of this two-stage method is first to segment individual teeth using \emph{i}MeshSegNet, an improvement of our MeshSegNet~\cite{Lian2020, Lian2019}.
After that, the cells belonging to each tooth (i.e., partial mesh) are cropped as an ROI for the localization of the corresponding landmarks.
That is, each ROI is fed into an individual PointNet-Reg, a variant of PointNet, to regress the heatmaps that encode the anatomical landmarks on the corresponding tooth.
By doing this, we narrow the possible locations of landmarks from the entire intraoral scan down to the specific ROIs, which significantly improves localization efficiency and accuracy.
%The results of the landmark positions from each ROI naturally locate on the original intraoral scan.

\begin{figure}[!t]
\centerline{\includegraphics[width=\columnwidth]{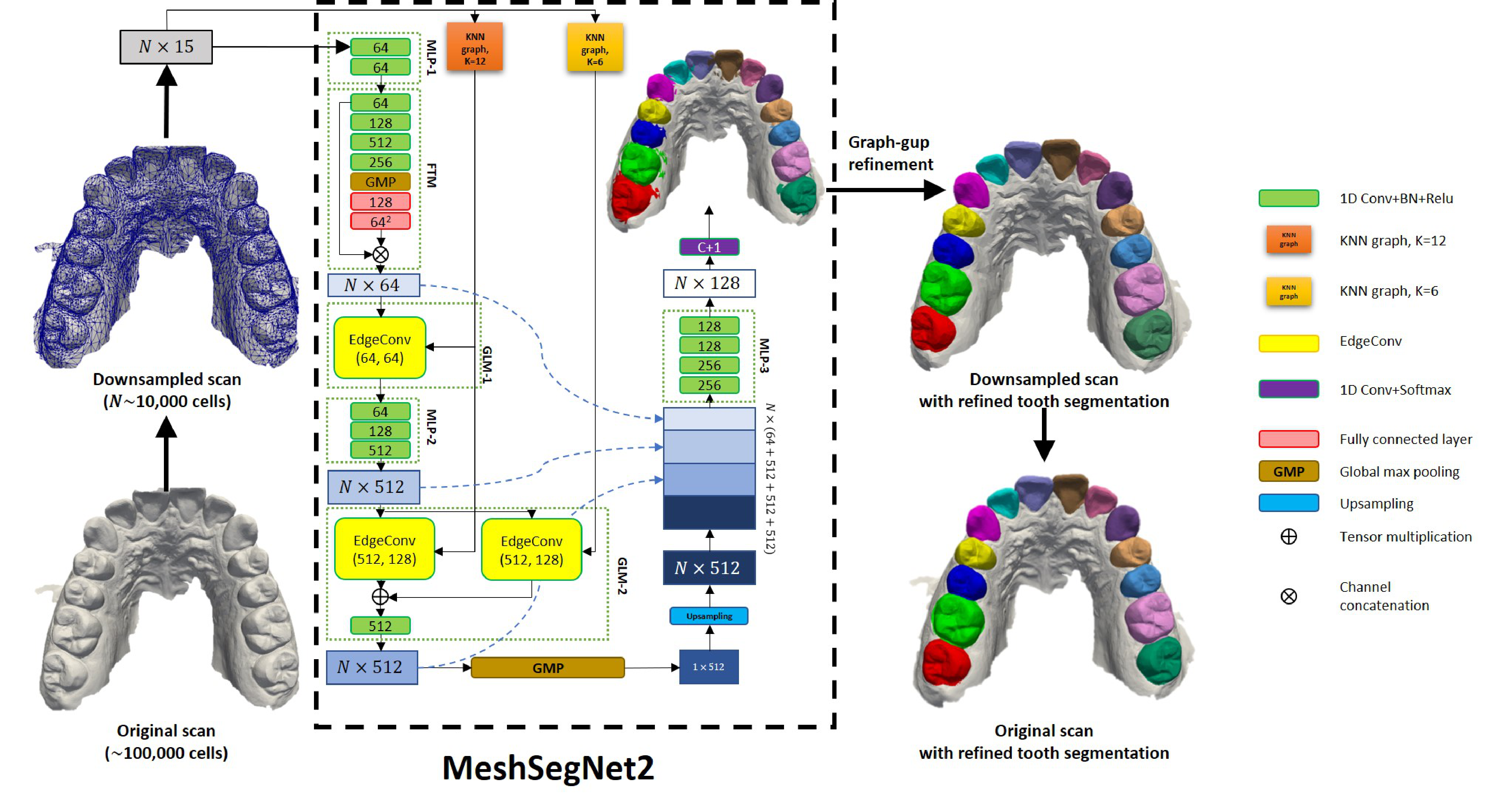}}
\caption{The network structure of \emph{i}MeshSegNet for tooth segmentation in Stage 1.}
\label{fig_imeshsegnet}
\end{figure}

\begin{figure}[!t]
\centerline{\includegraphics[width=\columnwidth]{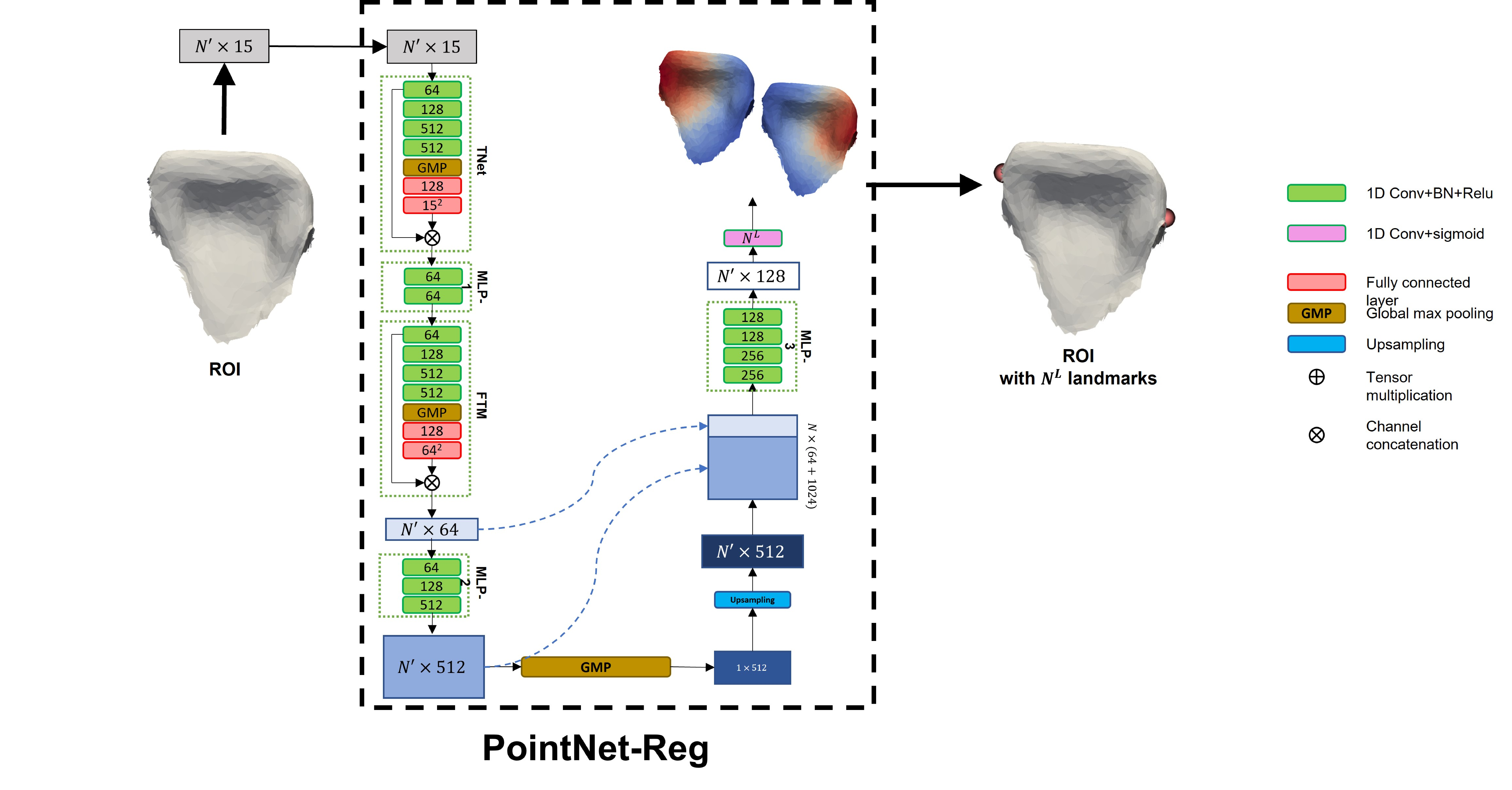}}
\caption{The network structure of PointNet-Reg for landmark regression in Stage 2.}
\label{fig_pointnet-reg}
\end{figure}

\subsubsection{\emph{i}MeshSegNet for Tooth Segmentation}
The purpose of Stage 1 is to perform automatic tooth segmentation on raw intraoral scans.
This stage includes three steps: pre-processing, inference, and post-processing.

\emph{In the pre-processing step}, the raw intraoral scans are first downsampled from \textcolor[rgb]{0,0,0}{approximately} 100,000 mesh cells (based on iTero Element\textregistered) to 10,000 cells.
The downsampled mesh further forms an $N\times15$ vector $\textbf{F}^0$, which is the input of \emph{i}MeshSegNet, where $N$ is \textcolor[rgb]{0,0,0}{the} number of cells. The $15$ dimensions correspond to coordinates of the three vertices of each cell (9 units), the normal vector of each cell (3 units), and the relative position of each cell with respect to the whole surface (3 units), respectively.
The $N\times15$ matrix is z-score normalized.

\emph{i}MeshSegNet inherits its forerunner, MeshSegNet~\cite{Lian2020, Lian2019}, and its architecture is shown in Fig.~\ref{fig_imeshsegnet}. The major difference between MeshSegNet and \emph{i}MeshSegNet is the implementation of the graph-constrained learning module (GLM) that has been proven effective for segmentation~\cite{Lian2020, Lian2019}. MeshSegNet utilizes symmetric average pooling (SAP) and two adjacency matrices ($A_S$ and $A_L$ in Refs.~\cite{Lian2020, Lian2019}) to extract local geometric contexts. However, the two $N\times N$ adjacency matrices and the matrix multiplication cause high computational complexity and substantial memory \textcolor[rgb]{0,0,0}{usage} when $N$ is large. In order to overcome this drawback, \emph{i}MeshSegNet adopts the EdgeConv operation~\cite{10.1145/3326362} to replace SAP for local context modeling.

\emph{In the inference step}, \emph{i}MeshSegNet first consumes the input $\textbf{F}^0$ ($N\times 15$) with a multi-layer perceptron module (MLP-1 in Fig.~\ref{fig_imeshsegnet}) to obtain an $N\times 64$ feature vector $\textbf{F}^1$.
A feature-transformer module (FTM) then predicts a $64\times 64$ transformation matrix $\textbf{T}$ based on the features learned by MLP-1, which maps the $\textbf{F}^1$ into a canonical feature space by performing a matrix (tensor) multiplication $\hat{\textbf{F}}^1 = \textbf{F}^1 \textbf{T}$.

The first graph-constrained learning module (i.e., GLM-1) applies EdgeConv~\cite{10.1145/3326362} to propagate the contextual information provided by neighboring cells on each centroid cell, resulting in a cell-wise feature matrix $\textbf{F}^{E1}=(f^{E1}_{1}, f^{E1}_{2}, ..., f^{E1}_{N})$ that encodes local geometric contexts, as expressed as
\begin{equation}
f^{E1}_{i} = \max_{j\in\mathfrak{N}(i)} h_{\Theta} \left(x_i-x_j, x_i \right),
\label{eq_1}
\end{equation}
where $j$ and $\mathfrak{N}(i)$ denote a single adjacent cell and all adjacent cells of cell $i$, computed by $k$-nearest neighbor ($k$-NN) graph; and $h_{\Theta}$ denotes a shared-weight MLP or a 1D Conv layer. Note that the $k$-NN graph is computed based on the initial graph including self-loop, which is unlike the dynamic graph update in~\cite{10.1145/3326362}, and we adopt $k=6$ for the $k$-NN graph in GLM-1.
The EdgeConv not only bypasses the expansive matrix multiplication but also has the permutation invariance and partial translation invariance properties~\cite{10.1145/3326362}, leading to better performance in terms of both efficiency and accuracy, as verified by our experiments presented in Section~\ref{sec_4}.

The feature matrix $\textbf{F}^{E1}$ then goes through the second MLP module (i.e., MLP-2), constituted by three shared-weight 1D Convs with 64, 128, 512 channels, respectively, becoming an $N\times 512$ feature matrix $\textbf{F}^2$ and passing through GLM-2.
Compared to GLM-1, GLM-2 further enlarges the receptive field to learn multi-scale contextual features. Specifically, $\textbf{F}^2$ is processed by two parallel EdgeConvs defined on two different $k$-NN graphs ($k=6$ and $k=12$), resulting in two feature matrices $\textbf{F}^{E2}$ and $\textbf{F}^{E2'}$.
They are then concatenated and fused by a 1D Conv with 512 channels.

A global max pooling (GMP) is applied on the output of GLM-2 to produce the translation-invariant holistic features that encode the semantic information of the whole dental mesh. Then, a dense fusion strategy is used to densely concatenate the local-to-global features from FTM, GLM-1, GLM-2, and upsampled GMP, followed by the third MLP module (MLP-3) to yield an $N\times 128$ feature matrix. Finally, a 1D Conv layer with softmax activation is used to predict an $N\times(C+1)$ probability matrix where $C=14$ (i.e., 14 teeth). The effectiveness of the GLM and dense fusion of local-to-global features have been examined systematically~\cite{Lian2020}.
All 1D Convs in \emph{i}MeshSegNet are followed by batch normalization (BN) and ReLU activation except the final 1D Conv.

\emph{In the post-processing step}, we first refine the segmentation results predicted by deep neural networks, as they may still contain isolated false predictions or non-smooth boundaries. Similar to~\cite{Xu2019, Lian2020}, we refine the network outputs by using the multi-label graph-cut method~\cite{Boykov2001} that optimizes
\begin{equation}
\operatorname*{argmin}_L \sum_{i=1}^{N} -log(p_i(l_i)+\epsilon) + \lambda \sum_{i=1}^{N} \sum_{j\in \mathscr{N}_i} V(p_i, p_j, l_i, l_{j}),
\label{eq_3}
\end{equation}
where the first and second terms are the data-fitting and local-smoothness terms, respectively; cell $i$ is labeled by the deep network with label $l_i$ under the probability of $p_i$; $\epsilon$ is a small scalar (i.e., $1\times 10^{-4}$) to ensure numerical stability; $\lambda$ is a non-negative tuning parameter; $j$ and $\mathscr{N}_i$ denote a single nearest-neighboring cell and all nearest-neighboring cells of cell $i$, respectively. The local smoothness term (the second term in Eq.~\ref{eq_3}) is
expressed as
\begin{equation}
V(p_i, p_j, l_i, l_{j}) =
\begin{cases}
    0,& l_i = l_j\\
		-log(\frac{\theta_{ij}}{\pi})\phi_{ij},& l_i \neq l_j, \text{$\theta_{ij}$ is concave}\\
    -\beta_{ij}log(\frac{\theta_{ij}}{\pi})\phi_{ij},& l_i \neq l_j, \text{$\theta_{ij}$ is convex}
\end{cases},
\label{eq_4}
\end{equation}
where $\theta_{ij}$ is the dihedral angle between cell $i$ and $j$; $\phi_{ij}= \left| c_i -c_j \right|$ and $\beta=1+\left| \hat{n}_i \cdot \hat{n}_j \right|$; $c$ and $\hat{n}$ denote the barycenter and normal vector of a cell, respectively. The term $\beta$ ($\beta=30$ in this study) enforces the optimization to favor concave regions as the boundaries among teeth and gingiva are usually concave~\cite{Xu2019}.

Finally, we project the segmentation result from the downsampled mesh (\textcolor[rgb]{0,0,0}{approximately} 10,000 cells) back to the original intraoral scan (\textcolor[rgb]{0,0,0}{approximately} 100,000 cells).
To this end, we train a support vector machine (SVM), with the radial basis function (RBF) kernel, using downsampled cells' coordinates and corresponding labels (i.e., predicted segmentations) as training data.
Then, we consider all cells in the original high-resolution intraoral scan as test data and predict their labels. By doing this, the prediction from the SVM model is the segmentation result under the original intraoral scan.

\subsubsection{Point-Reg for Landmark Localization}\label{sec_stage-2}
A tooth anatomical landmark locates on and is determined by the shape of its related tooth.
For example, both the mesial and distal contact points (i.e., MCP and DCP in Table~\ref{table_1}) defined in this study refer to the center of contact areas which are usually located in the upper one third of the crown on the mesial and distal sides of most teeth. Therefore, it is reasonable and more efficient to locate a landmark only using the mesh of the corresponding tooth, instead of the entire dental arch model.

Then, we modify the original PointNet~\cite{Qi2016} for cell-wise regression, called PointNet-Reg, which learns the non-linear mapping from a tooth ROI to the corresponding heatmaps that encode landmark locations.
As shown in Fig.~\ref{fig_pointnet-reg}, we replace the softmax activation function to the sigmoid function in the final 1D Conv to output an $N'\times N^L$ Gaussian heatmap matrix that encodes \textcolor[rgb]{0,0,0}{the} location of the landmarks, where $N'$ and $N^{L}$ represent the number of cells in the ROI and the number of anatomical landmarks, respectively.
After the inference by PointNet-Reg, we choose the centroid cells having the maximum heatmap values as the predicted landmark coordinates, which naturally locate on the original intraoral scan.

\subsection{Implementation}\label{implementation}
\subsubsection{Data Augmentation}\label{sec_augmentation}
Due to the symmetry of the dental arch, all \textcolor[rgb]{0,0,0}{136} samples with their annotated labels as well as \textcolor[rgb]{0,0,0}{66} landmarks are flipped to double the sample size.
Then, we follow the same strategy proposed in~\cite{Lian2020, Lian2019} to augment these original and flipped data, by combining \textcolor[rgb]{0,0,0}{the} operations of 1) random translation, 2) random rotation, and 3) random rescaling. Specifically, along each axis in the 3D space, an intraoral scan has 50\% probability to be translated between [-10, 10], rotated between [-$\pi$, $\pi$], and scaled between [0.8, 1.2], respectively. The combination of these random operations generates 20 "new" cases from each original and flipped scan in this study.

\subsubsection{Heatmap Generation}
The ground-truth heatmaps in PointNet-Reg are defined by a Gaussian function, expressed as
\begin{equation}
h_k(\textbf{x}_i) = H\exp(-\frac{(\textbf{x}_{i}-\textbf{x}^{L}_k)^2}{2\sigma_H^2}),
\label{eq_5}
\end{equation}
where $\textbf{x}_i$ and $\textbf{x}^L_k$ are the coordinates of centroid cell $i$ and landmark $k$, respectively; $h_k$ and $H$ are the Gaussian height for landmark $k$ and the maximum Gaussian height ($H=1.0$); and $\sigma_H$ is the Gaussian root mean square (RMS) width ($\sigma_H=5.0\, mm$).

\subsubsection{Training Procedure}\label{training_procedure}
The model was trained by minimizing the generalized Dice loss~\cite{Sudre2017} in Stage 1 (i.e., \emph{i}MeshSegNet for tooth segmentation) and mean square error (MSE) in Stage 2 (i.e., PointNet-Reg for landmark regression), respectively, using the AMS-Grad variant of the Adam optimizer~\cite{Reddi2019}.
In the training step, we randomly selected 9,000 and \textcolor[rgb]{0,0,0}{3,000} cells from the downsampled scan (approximate 10,000 cells) and \textcolor[rgb]{0,0,0}{ROI in the original resolution} (approximate 3,050 cells) to form the inputs of \emph{i}MeshSegNet (i.e., $N=9,000$) in Stage 1 and PointNet-Reg (i.e., $N'=\textcolor[rgb]{0,0,0}{3,000}$) in Stage 2, respectively.

\subsubsection{Inference Procedure}
The proposed TS-MDL can process the intraoral \textcolor[rgb]{0,0,0}{scans of varying sizes}.
We downsample the unseen intraoral scan before applying the trained \emph{i}MeshSegNet model on it, which is due to the limited GPU memory (e.g., 11GB for NIVIDIA RTX 2080 Ti).
At the end of Stage 1, we transfer the downsampled segmentation result predicted by \emph{i}MeshSegNet back to its original resolution.
Then, \textcolor[rgb]{0,0,0}{14} ROIs are generated in Stage 2, and each ROI is fed into the corresponding PointNet-Reg under the original resolution.
In total, \textcolor[rgb]{0,0,0}{66} heatmaps are predicted from \textcolor[rgb]{0,0,0}{14} different PointNet-Reg models and converted to the corresponding landmark coordinates.
In this study, we conducted experiments to localize \textcolor[rgb]{0,0,0}{66} landmarks on \textcolor[rgb]{0,0,0}{14} teeth from the upper dental model, but our TS-MDL method could be straightforwardly extended to include more landmarks and teeth \textcolor[rgb]{0,0,0}{as well as lower dental models (with in-house tests and data not shown)} if needed.

\section{Experiments}
\label{sec_4}

\subsection{Experimental Setup}
\textcolor[rgb]{0,0,0}{We split the 136 samples into training, validation, and test sets with a ratio of 65:15:20.}
Both training and validation sets were augmented using the method described in Sec.~\ref{sec_augmentation}.
Using the manual annotations as the ground truth, the segmentation performance in Stage 1 was quantitatively evaluated by Dice similarity coefficient (DSC), sensitiviy (SEN), postive predictive value (PPV), \textcolor[rgb]{0,0,0}{and Hausdorff distance (HD)}, and the landmark localization performance in Stage 2 was evaluated by the mean absolute error (MAE), which is defined as
\begin{equation}
\text{MAE}(\textbf{x}^P_{k}, \textbf{x}^L_{k}) = \frac{1}{N^L}\sum^{N^L}_{k=1} \left| \textbf{x}^P_{k} - \textbf{x}^L_{k} \right|,
\label{eq_6}
\end{equation}
where $\textbf{x}^P_{k}$ and $\textbf{x}^L_{k}$ denote the predicted and ground-truth coordinates of landmark $k$, respectively; $N^{L}$ represents the number of anatomical landmarks.

\subsection{Competing Methods}
In the task of tooth segmentation, we compare \emph{i}MeshSegNet with its forerunner MeshSegNet.
Note that in our previous work~\cite{Lian2020, Lian2019}, MeshSegNet has already been compared with other state-of-the-art methods, showing superior performance.

In the task of landmark identification, our \textbf{\emph{i}MeshSegNet+PointNet-Reg} is compared with another two-stage method (i.e., \textbf{\emph{i}MeshSegNet+MeshSegNet-Reg}), where PointNet-Reg is replaced by MeshSegNet-Reg (a variant of MeshSegNet) to regress landmark heatmaps for each tooth based on its segmentation produced by \emph{i}MeshSegNet.
In addition, to verify if tooth segmentation can boost landmark localization, these two-stage methods are also compared with two single-stage methods (i.e., \textbf{PointNet-Reg} and \textbf{\emph{i}MeshSegNet-Reg}), which regress landmark heatmaps in an end-to-end fashion from the input 3D dental model.

\subsection{Results}
\subsubsection{Tooth Segmentation Results}
\begin{figure}[!t]
\centerline{\includegraphics[width=\columnwidth]{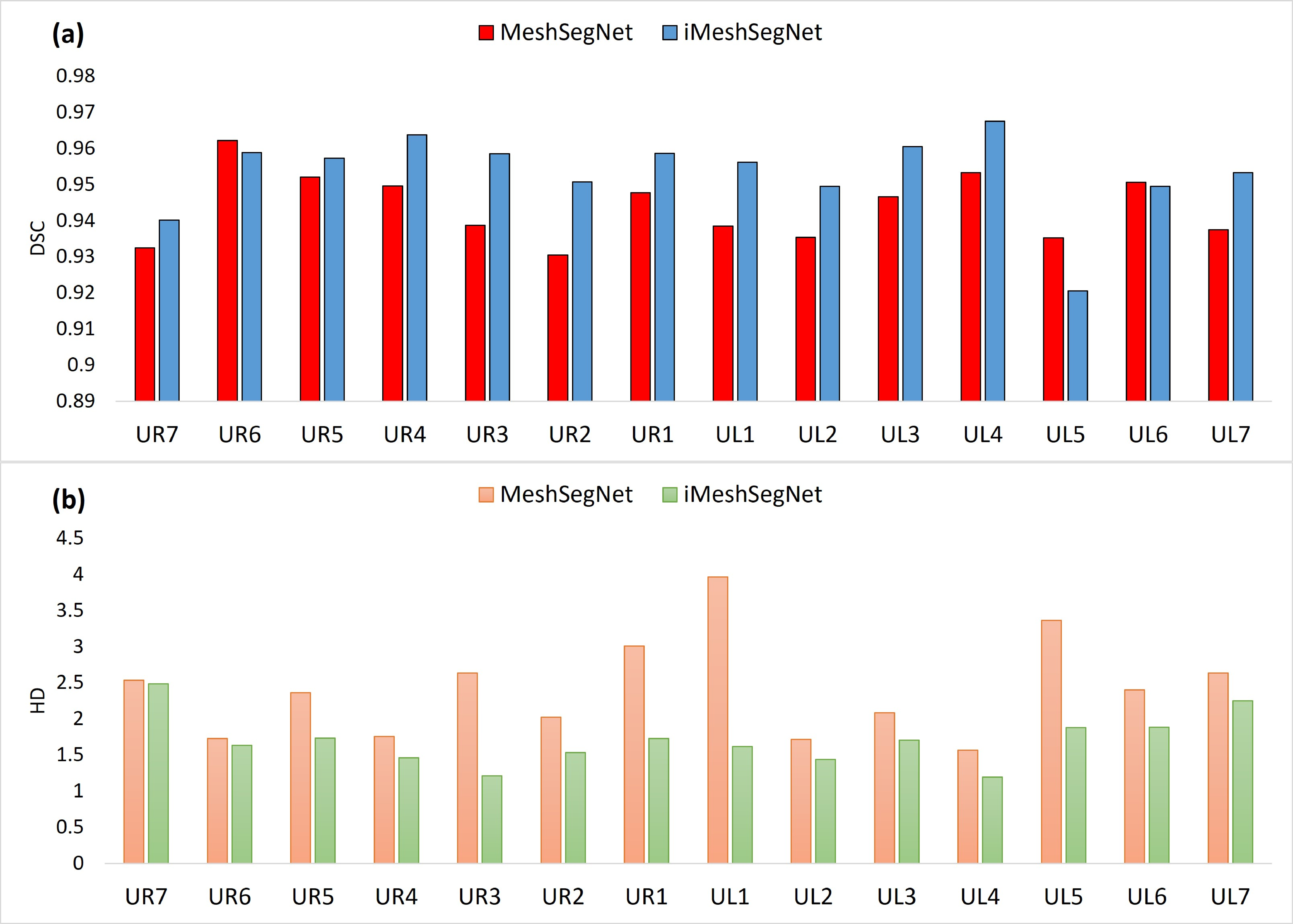}}
\caption{The results of \textcolor[rgb]{0,0,0}{(a) Dice similarity coefficient (DSC) and (b) Hausdorff distance (HD)} of each tooth from MeshSegNet and \emph{i}MeshSegNet.}
\label{fig_DSC_comparison}
\end{figure}

The quantitative segmentation results obtained by MeshSegNet and \emph{i}MeshSegNet, both without post-processing, are compared in Table~\ref{table_2}, which show that \emph{i}MeshSegNet significantly outperformed the original MeshSegNet in terms of all \textcolor[rgb]{0,0,0}{four} metrics (i.e., DSC, SEN, PPV, and \textcolor[rgb]{0,0,0}{HD}).
Fig.~\ref{fig_DSC_comparison} further summarizes the segmentation result (in terms of DSC and \textcolor[rgb]{0,0,0}{HD}) for each tooth, from which we can see that, being consistent with the average segmentation accuracy in Table~\ref{table_2}, \emph{i}MeshSegNet also led to better accuracy in segmenting each tooth.
\textcolor[rgb]{0,0,0}{The \emph{i}MeshSegNet segmentation results refined by the multi-label graph cut are DSC of $0.964 \pm 0.054$, SEN of $0.970 \pm 0.061$, PPV of $0.960 \pm 0.054$, and HD of $0.995 \pm 0.722$ $mm$, respectively.
}

\begin{figure}[!t]
\centerline{\includegraphics[width=\columnwidth]{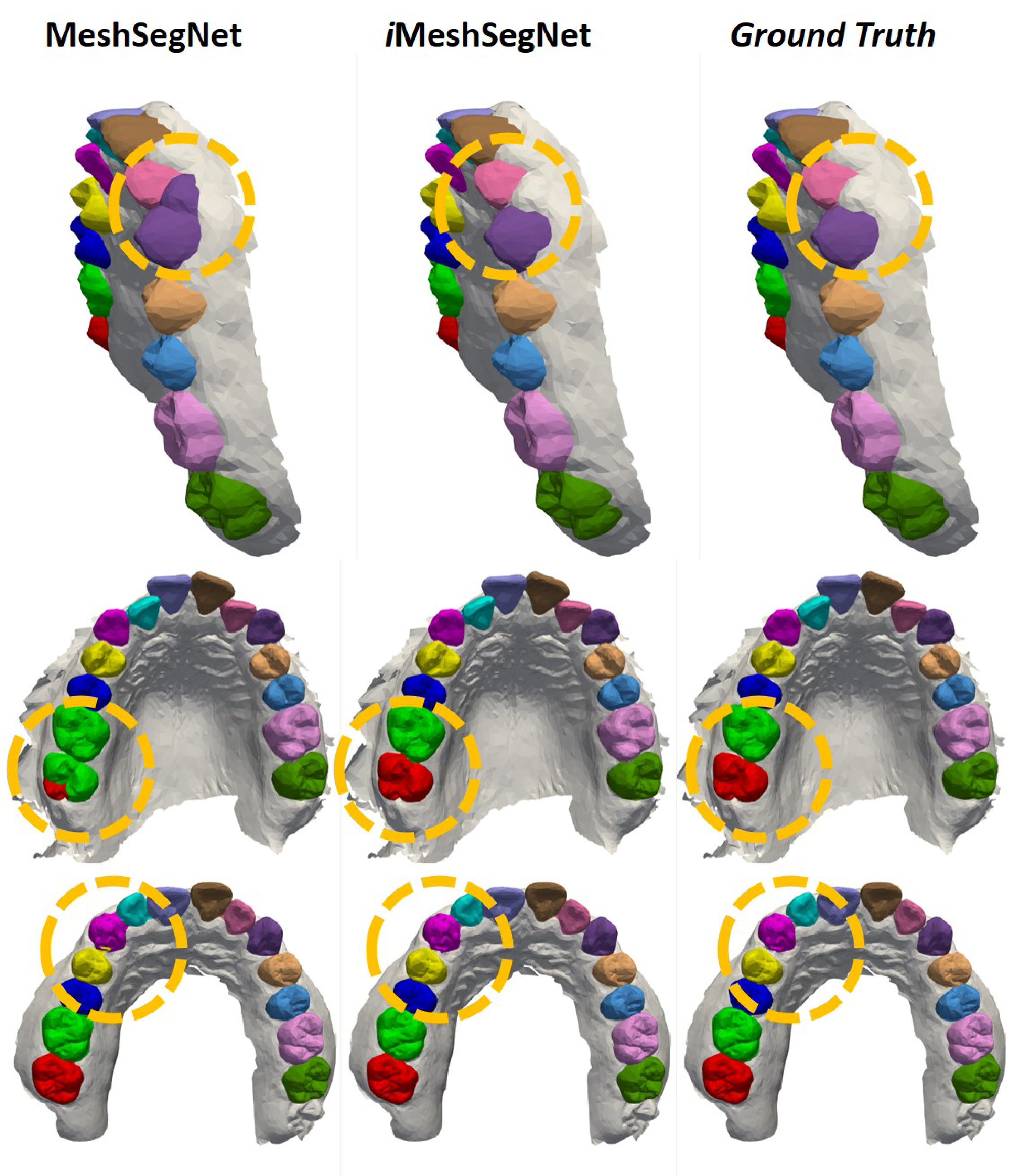}}
\caption{The qualitative comparison among {\emph{i}MeshSegNet}, MeshSegNet, and ground truth. Each row contains the automated segmentation results of a sample labeled by different methods (with post-processing refinement). The yellow dotted circles highlight the areas that \emph{i}MeshSegNet \textcolor[rgb]{0,0,0}{outperformed} its forerunner MeshSegNet, compared to the ground turth.}
\label{fig_visual_segmentation_comparison}
\end{figure}

A qualitative comparison among {\emph{i}MeshSegNet}, MeshSegNet, and the ground truth is visualized in Fig.~\ref{fig_visual_segmentation_comparison}.
Each row presents the segmentations of a representative example produced by the two automatic methods (with post-processing refinement) and the ground truth, respectively. 
We can see that \emph{i}MeshSegNet outperformed its forerunner MeshSegNet in the challenging areas (e.g., those highlighted by the yellow dotted circles).

\begin{table}[]
\centering
\caption{The comparison of segmentation results without multi-label graph cut refinement between MeshSegNet and \emph{i}MeshSegNet in terms of Dice similarity coefficient (DSC), sensitivity (SEN), positive predictive value (PPV)\textcolor[rgb]{0,0,0}{, and Hausdorff distance (HD)}. Bold font indicates the best result.}
\begin{tabular}{lcc}
\hline
Metric & \multicolumn{1}{l}{MeshSegNet~\cite{Lian2020, Lian2019}}                                           & \textbf{\emph{i}MeshSegNet} \\ \hline
DSC    & \begin{tabular}[c]{@{}c@{}}\textcolor[rgb]{0,0,0}{0.943 $\pm$ 0.044}\\ \textcolor[rgb]{0,0,0}{\textit{p}=0.0006}\end{tabular} & \textcolor[rgb]{0,0,0}{\textbf{0.953 $\pm$ 0.056}} \\
SEN    & \begin{tabular}[c]{@{}c@{}}\textcolor[rgb]{0,0,0}{0.946 $\pm$ 0.054}\\ \textcolor[rgb]{0,0,0}{\textit{p}=0.0183}\end{tabular}           & \textcolor[rgb]{0,0,0}{\textbf{0.955 $\pm$ 0.064}} \\
PPV    & \begin{tabular}[c]{@{}c@{}}\textcolor[rgb]{0,0,0}{0.943 $\pm$ 0.051}\\ \textcolor[rgb]{0,0,0}{\textit{p}=0.0014}\end{tabular} & \textcolor[rgb]{0,0,0}{\textbf{0.953 $\pm$ 0.058}} \\
HD    & \begin{tabular}[c]{@{}c@{}}\textcolor[rgb]{0,0,0}{2.420 $\pm$ 3.140 $mm$}\\ \textcolor[rgb]{0,0,0}{\textit{p}=0.0209}\end{tabular} & \textcolor[rgb]{0,0,0}{\textbf{1.696 $\pm$ 1.087 $mm$}} \\
\hline
\end{tabular}
\label{table_2}
\end{table}

Besides the segmentation accuracy, we further compared the computational efficiency between MeshSegNet and \emph{i}MeshSegNet. The training and inference time of these two methods, using the same implementation strategy and computational environment, are shown in Table~\ref{table_3}.
From Table~\ref{table_3}, we can observe that the average training time of \emph{i}MeshSegNet is $253.6\, sec$ per epoch \textcolor[rgb]{0,0,0}{for 30 samples}, roughly 20 times faster than MeshSegNet.
In addition, \emph{i}MeshSegNet only needs roughly $0.62\, sec$ to conduct the segmentation of an unseen input, which is approximately 8.6 times faster than MeshSegNet.

The above results show that \emph{i}MeshSegNet significantly outperformed the original MeshSegNet in both accuracy and efficiency, which implies the efficacy of the substitution of SAP and adjacency matrices by EdgeConv.

\begin{table}[]
\centering
\caption{The comparison of computing time between MeshSegnet and \emph{i}MeshSegNet. Bold font indicates the best result.}
\begin{tabular}{lcc}
\hline
Metric & \multicolumn{1}{l}{MeshSegNet~\cite{Lian2020, Lian2019}}                                           & \textbf{\emph{i}MeshSegNet} \\ \hline
Training (sec/epoch)    & 5114.4 & \textbf{253.6}\\
Prediction (sec/scan)    & 5.33 & \textbf{0.62} \\ \hline
\end{tabular}
\label{table_3}
\end{table}

\subsubsection{Landmark Localization Results}
The results of the four competing methods in terms of MAE are summarized in Table~\ref{table_4}, which lead to three observations.
First, the two-stage methods (i.e, \emph{i}MeshSegNet+PointNet-Reg and \emph{i}MeshSegNet+\emph{i}MeshSegNet-Reg) consistently outperformed the single-stage methods, demonstrating the efficacy of the use of ROI.
Second, the single-stage \emph{i}MeshSegNet-Reg shows a better MAE of \textcolor[rgb]{0,0,0}{$1.566\pm 3.711\,mm$} compared to PointNet-Reg.
This finding matches our expectations since \emph{i}MeshSegNet incorporates a series of GLMs and a dense fusion strategy to learn higher-level features from the raw intraoral scans.
Third, \emph{i}MeshSegNet+PointNet-Reg obtained better results than \emph{i}MeshSegNet+\emph{i}MeshSegNet-Reg.
Two possible explanations for this controversial finding might be that 1) the ROI implicitly provides the localized information (i.e., the learning difficulty has been significantly reduced) so the advantage of \emph{i}MeshSegNet-Reg on extracting localized information (i.e., knowing where it is in the whole scan) is not as important as in the single-stage, and 2) due to the reduced learning difficulty in the two-stage manner and our small dataset, \emph{i}MeshSegNet-Reg might slightly overfit and its generalization ability is worse on the test set, compared to PointNet-Reg.

\begin{table}
\centering
\caption{The results of the four competing methods (two single-stage and two two-stage strategies) in terms of mean absolute error (MAE) for the landmark localization. Bold font indicates the best result.}
\begin{tabular}{ll}
\hline
Method                                     & MAE (mm)              \\
\hline
1-stage: PointNet-Reg                      & \textcolor[rgb]{0,0,0}{1.807 $\pm$ 1.558}           \\
1-stage: \emph{i}MeshSegNet-Reg                   & \textcolor[rgb]{0,0,0}{1.250 $\pm$ 1.021}           \\
\textbf{2-stage: \emph{i}MeshSegNet+PointNet-Reg} & \textcolor[rgb]{0,0,0}{\textbf{0.597 $\pm$ 0.761}}  \\
2-stage: \emph{i}MeshSegNet+\emph{i}MeshSegNet-Reg       & \textcolor[rgb]{0,0,0}{0.773 $\pm$ 0.832}           \\
\hline
\end{tabular}
\label{table_4}
\end{table}

\begin{figure}[!t]
\centerline{\includegraphics[width=\columnwidth]{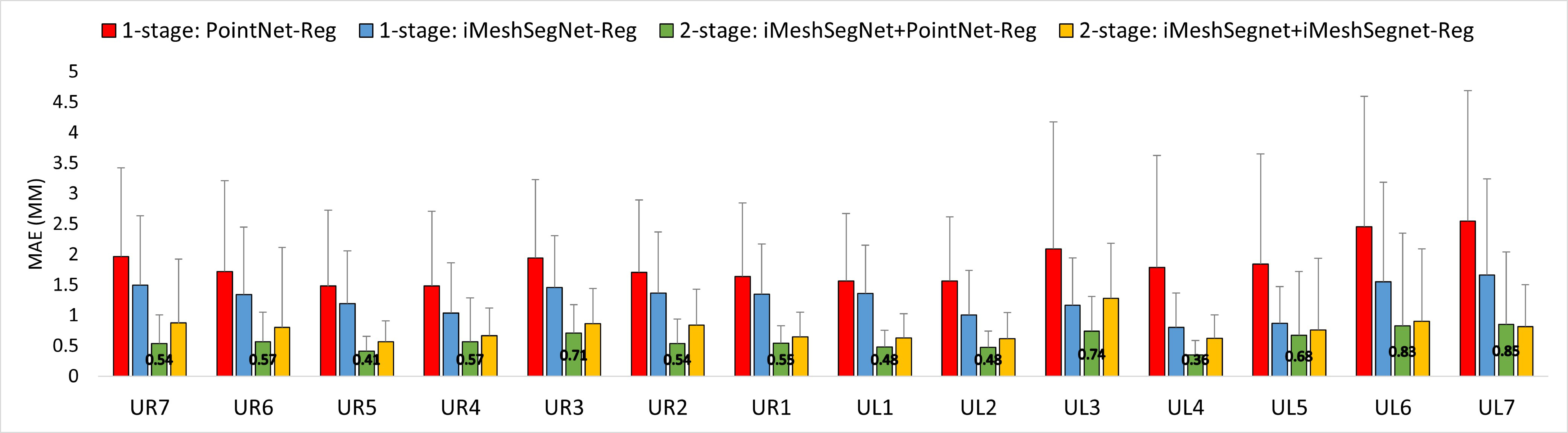}}
\caption{The mean absolute error (MAE) of landmarks on each tooth predicted by the four competing methods. The numbers showing on the green bars are the MAE predicted by \emph{i}MeshSegNet+PointNet-Reg for each tooth.}
\label{fig_MAE_comparison}
\end{figure}

Fig.~\ref{fig_MAE_comparison} displays the MAE of landmarks on each tooth predicted by the four competing methods. The results from each tooth have similar trends as the overall evaluation reported in Table~\ref{table_4}.
Although there is no public dataset of intraoral scans and benchmark of mesh segmentation for fair comparison, it is worth noting that our results achieve comparable accuracies, even slightly better, compared to Ref.~\cite{Sun2020a}.
The MAEs for the incisor, canine, premolar, and molar are \textcolor[rgb]{0,0,0}{$0.51\, mm$, $0.72\, mm$, $0.51\, mm$, and $0.70\, mm$}, respectively, from our \emph{i}MeshSegNet+PointNet-Reg, whereas the MAEs for the same categories are $0.65\,mm$, $0.71\, mm$, $0.86\, mm$, and $0.96\, mm$, respectively, in Ref.~\cite{Sun2020a}.
Moreover, higher errors of landmark localization are observed on the first molar (i.e., UR6 and UL6), which is similar to the trend in the segmentation task that prediction accuracy of the molar (UR/L6, UR/L7) is lower than the other teeth.
One possible reason is that uncompleted capture often occurs on the molars due to the difficulty in scanning the posterior areas, resulting in the lack of complete information on molars.

\begin{figure}[!t]
\centerline{\includegraphics[width=\columnwidth]{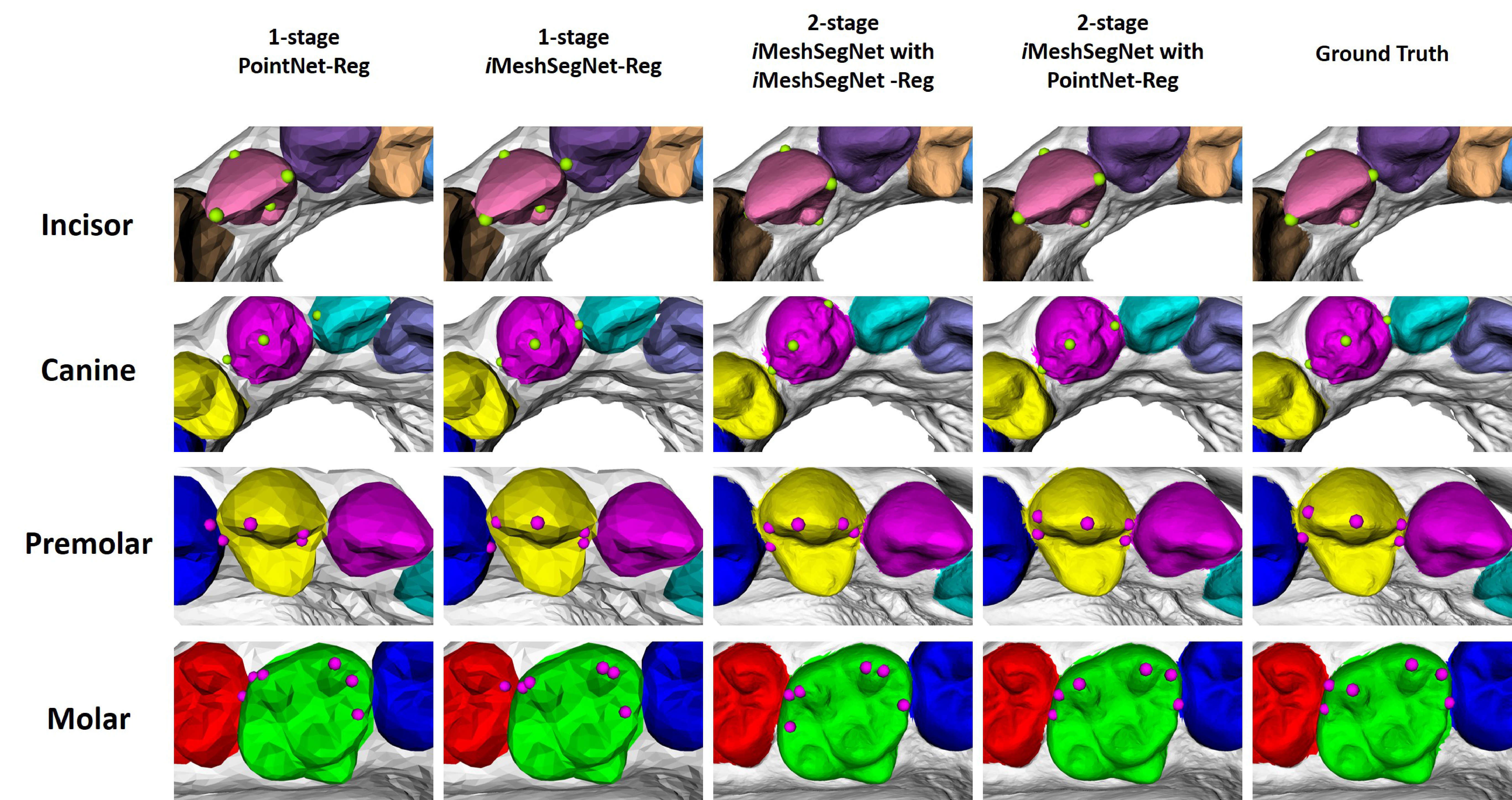}}
\caption{The qualitative comparison among the four competing methods and ground truth for different teeth. The green and purple circles represent landmarks in anterior teeth (i.e., incisor and canine) and posterior teeth (i.e., premolar and molar), respectively. The tooth color also indicates its segmentation result. Due to the limited GPU memory, both one-stage strategies (the first and second columns) predict results under downsampled meshes.}
\label{fig_visual_landmark_comparison}
\end{figure}

Furthermore, Fig.~\ref{fig_visual_landmark_comparison} illustrates the qualitative comparisons among the four competing methods and the ground truth for different teeth.
The green and purple circles represent landmarks in anterior teeth (i.e., incisor and canine) and posterior teeth (i.e., premolar and molar), respectively. The qualitative observation reveals that \emph{i}MeshSegNet+PointNet-Reg (the fourth column) has the best prediction, which is consistent with the quantitative measurement given in Fig.~\ref{fig_MAE_comparison}.

\section{Discussion}

\subsection{Static Adjacency vs. Dynamic Adjacency}
Wang et al.~\cite{10.1145/3326362} suggest using dynamic adjacency matrices defined in high-dimensional feature spaces to model the non-local dependencies between points (or cells) along the forward path of the network.
However, due to the specific geometric distribution and arrangement of human teeth, we think it is more reasonable to use the static adjacency (defined in the input Euclidean space) when extracting contextual features even in deeper layers.
To justify this claim, we compared the tooth segmentation results obtained by the aformentioned strategies, as shown in Table~\ref{table_5}.
From Table~\ref{table_5}, we can see that the model using static adjacency surpassed the one using dynamic adjacency in terms of DSC, SEN, and PPV in our specific task of tooth segmentation on 3D intraoral scans.

\begin{table}[]
\centering
\caption{The comparison of segmentation results based on static adjacency and dynamic adjacency. Bold font indicates the best result.}
\begin{tabular}{lcl}
\hline
Metric & Dynamic adjacency                                           & Static adjacency \\ \hline
DSC    & \begin{tabular}[c]{@{}c@{}}\textcolor[rgb]{0,0,0}{0.858 $\pm$ 0.169}\\ \textit{p}\textless{}0.0001\end{tabular} & \textcolor[rgb]{0,0,0}{\textbf{0.953 $\pm$ 0.056}} \\
SEN    & \begin{tabular}[c]{@{}c@{}}\textcolor[rgb]{0,0,0}{0.900 $\pm$ 0.175}\\ \textcolor[rgb]{0,0,0}{\textit{p}=0.0027}\end{tabular} & \textcolor[rgb]{0,0,0}{\textbf{0.955 $\pm$ 0.064}} \\
PPV    & \begin{tabular}[c]{@{}c@{}}\textcolor[rgb]{0,0,0}{0.834 $\pm$ 0.165}\\ \textit{p}\textless{}0.0001\end{tabular} & \textcolor[rgb]{0,0,0}{\textbf{0.953 $\pm$ 0.058}} \\ \hline
\end{tabular}
\label{table_5}
\end{table}

\begin{table}[]
\centering
\caption{Ceiling analysis of our TS-MDL.}
\begin{tabular}{ccc}
\hline
\textbf{Stage} & \textbf{Accuracy (MAE, mm)} & \textbf{Improvement (mm)} \\ \hline
Overall        & \textcolor[rgb]{0,0,0}{$0.597\pm 0.761$}                      & N/A                       \\
Stage 1        & \textcolor[rgb]{0,0,0}{$0.510\pm 0.477$}                      & \textcolor[rgb]{0,0,0}{0.087}                     \\
%\textcolor[rgb]{0,0,0}{Stage 2}        & $0.112\pm 0.161$                      & 0.388                      \\ \hline
\end{tabular}
\label{table_7}
\end{table}

\subsection{Clinically Acceptable Errors}
\textcolor[rgb]{0,0,0}{According to Tanikawa et. al.~\cite{Tanikawa2009}, several methods were used to evaluate the clinically acceptable range in orthodontics. The simplest computational method is to determine whether the predicted landmarks are located in a circle within a 2 $mm$ radius.}
\textcolor[rgb]{0,0,0}{In addition, the American Board of Orthodontics (ABO) objective grading system considers deviations of 0.5 $mm$ as an awareness distinction~\cite{Casko1998}, implying 0.5 $mm$ is a high-standard criterion. The deviations over 0.5 $mm$ and 1 $mm$ will be penalized 1 point and 2 points, respectively, for the alignment and marginal ridge categories.}
\textcolor[rgb]{0,0,0}{Based on these definitions, we can see that the results predicted from our TS-MDL in Fig.~\ref{fig_MAE_comparison} meet the criterion reported in Ref.~\cite{Tanikawa2009} and approximate the marginal ridge and alignment criteria for the ABO.}
\textcolor[rgb]{0,0,0}{However, a ``passing`` case usually has an ABO score of 27 points or less, indicating that errors greater than 0.5 $mm$ can still achieve clinically acceptable results.}

\subsection{Ceiling Analysis}
We performed a ceiling analysis, where the ground truth of segmentation was used as input in Stage 1.
The analysis is given in Table~\ref{table_7}.
Perfect segmentation resulted in $\textcolor[rgb]{0,0,0}{0.087}\, mm$ improvement in overall accuracy in terms of MAE in the proposed TS-MDL.
Based on the ceiling analysis, it appears that refining the work of landmark regression in the future will yield the best overall improvement in the similar multi-stage manner\textcolor[rgb]{0,0,0}{, particularly if we consider 0.5 $mm$ as the requirement in terms of accuracy}.

\subsection{Limitations and Future Work}
Although our TS-MDL leverages the ROI to achieve state-of-the-art performance, it still has some limitations.
First, the dataset  only contains \textcolor[rgb]{0,0,0}{136} samples, which is relatively small.
In the future, we will keep collecting intraoral scans used in dental clinics.
Second, intraoral scans only have surface information. In the heatmap regression-based method, we can only predict those landmarks on the dental mesh surface.
However, due to areas of malocclusion, some teeth are overlapped with adjacent teeth, resulting in an incomplete mesh.
If a landmark happens to be located on the overlapped area, then the heatmap regression-based method is unable to accurately predict its location.
To solve this issue, one of our future works is to introduce 3D dental mesh repairing at the end of Stage 1 in order to reconstruct the incomplete areas from intraoral scans.
% Alternatively, to develop a regression network that can predict landmarks in the air instead of on the surface only.

\section{Conclusion}
This study has a two-fold contribution. First, we proposed an end-to-end graph-based neural network, \emph{i}MeshSegNet, for automated tooth segmentation on dental intraoral \textcolor[rgb]{0,0,0}{scans}, which improves upon our previous work in terms of the implementation of graph-constrained learning \textcolor[rgb]{0,0,0}{modules}. \emph{i}MeshSegNet shows significantly better accuracy in terms of DSC, SEN, PPV\textcolor[rgb]{0,0,0}{, and HD} and dramatically reduces computational time in both training and prediction.
Second, we proposed the TS-MDL to automatically localize tooth landmarks on intraoral scans. In TS-MDL, we first predict the segmentation masks using \emph{i}MeshSegNet with graph-cut refinement. We consider the segmentation masks as independent ROIs and then feed them into a series of regression network PointNet-Reg to predict the heatmaps that encode the coordinates of tooth landmarks. Our method can achieve \textcolor[rgb]{0,0,0}{an averaged} MAE of $\textcolor[rgb]{0,0,0}{0.597\pm 0.761}\, mm$ for localizing \textcolor[rgb]{0,0,0}{66} landmarks, showing that it \textcolor[rgb]{0,0,0}{has the potential to} be \textcolor[rgb]{0,0,0}{utilized in orthodontic applications}.

%\appendices
%
%Appendixes, if needed, appear before the acknowledgment.

\section*{Acknowledgment}
This work is supported, in part, by the Ohio State University College of Dentistry, NIH\/NIDCR DE022816, and NSF\#1938533. We also gratefully acknowledge the support of computing resource provided by the Ohio Supercomputer Center.

\section*{Disclosure statement}
Dr. Ko is the co-funder of SOVE Inc. However, the work presented here has no financial involvement with SOVE.

\bibliographystyle{ieeetran}
\bibliography{TMI_Wu_2020}{}

\end{document}